\documentclass[preprint,12pt]{elsarticle}

\usepackage{amssymb}
\usepackage{lmodern}
\usepackage{amsmath}
\usepackage{graphicx}
\usepackage{booktabs}
\usepackage{tabularx}
\usepackage{array}
\usepackage{url}
\usepackage{placeins}
\usepackage{algorithm}
\usepackage{algpseudocode}
\usepackage{multirow}
\usepackage{microtype}
\usepackage[hidelinks]{hyperref}
\usepackage{bookmark}
\hypersetup{pdfauthor={Rushab Rasik Karania and Tomas Maul},
  pdftitle={Query-Conditioned Prototype Adaptation for Cross-Domain Few-Shot Learning}}


\journal{Neurocomputing}

\begin{document}

\begin{frontmatter}

\title{Query-Conditioned Prototype Adaptation for Cross-Domain Few-Shot Learning:
Single-Query Inference, Controlled Comparisons, and Failure Modes}

\author[unm]{Rushab Rasik Karania}
\ead{rushabkarania@gmail.com}

\author[unm]{Tomas Maul\corref{cor1}}
\ead{tomas.maul@nottingham.edu.my}
\cortext[cor1]{Corresponding author.}

\affiliation[unm]{organization={School of Computer Science, University of Nottingham Malaysia},
            city={Semenyih},
            postcode={43500},
            state={Selangor},
            country={Malaysia}}

\begin{abstract}
Cross-domain few-shot learning requires adapting a classifier to a new visual
domain from very few labelled examples without target-time parameter updates.
We isolate one question: under a fixed global representation, what does joint
query--support adaptation contribute to prototype construction? The
Within-Instance Prototypical Transformer (WIPT) implements single-query
test-time prototype adaptation by jointly transforming one unlabelled query and
the labelled support embeddings, then forming query-specific class means.
Using a shared frozen ViT-S/16 encoder, miniImageNet source training, and CUB,
EuroSAT and ISIC targets, we replicate the key comparisons across five
independent training seeds. In 1-shot evaluation, WIPT improves frozen ProtoNet
in every run on CUB ($+0.21$ percentage points) and EuroSAT ($+2.07$), but
decreases ISIC ($-0.22$). In 5-shot evaluation, ProtoNet remains strongest
overall, while WIPT consistently improves a capacity-matched support-only
Transformer on ISIC ($+0.99$). Joint processing of up to five queries yields no
reliable accuracy gain; in a head-only 5-shot benchmark, $g=5$ reduces
analytical attention-token pairs by 73\% and peak allocated memory by 29\%
relative to $g=1$, although latency is non-monotonic. Across all target/shot
conditions, WIPT changes uncertain ProtoNet decisions far more than confident
ones, and rescue/break decomposition accounts for the observed gains and
losses. Source-shift and scorer controls further show that the benefit is not
universal. Overall, WIPT provides a streaming-compatible form of test-time
prototype adaptation that can improve difficult low-shot cross-domain decisions
without target-time optimization.

\end{abstract}


\end{frontmatter}

\section{Introduction}
\label{sec:intro}
Cross-domain few-shot learning (CD-FSL) asks a model trained on a source domain
to recognize new target-domain classes from very few labelled examples
\cite{tseng2020fwt,guo2020bscdfsl}. We study this problem under a fixed image
representation: adaptation is restricted to the episodic classification head.
ProtoNet forms a class representative by averaging its support embeddings
\cite{snell2017prototypical}. With limited support, this fixed mean may be an
imperfect reference for a particular query.

The Within-Instance Prototypical Transformer (WIPT) jointly transforms one
unlabelled query and the labelled support embeddings, then averages the
transformed supports class-wise. Both the query representation and its class
prototypes depend on this interaction. The encoder stays frozen, with no
target-time gradient update or pseudo-labelling. Query-aware adaptation is
established in prior work; our contribution is a controlled investigation of
joint query participation using one global embedding per image.

We compare WIPT with frozen ProtoNet and a capacity-matched support-only
Transformer across five independent training seeds, three target datasets and
two shot regimes. Four questions organize the study:
\begin{enumerate}
\item[\textbf{RQ1}] When does joint query participation improve over fixed
prototypes and the matched support-only transformation?
\item[\textbf{RQ2}] Does processing multiple queries together improve accuracy,
and what computational trade-offs result?
\item[\textbf{RQ3}] How do decision changes vary with domain, shot count and
proximity to the ProtoNet boundary?
\item[\textbf{RQ4}] Does source-only pseudo-domain training improve transfer?
\end{enumerate}

The findings are conditional. WIPT improves ProtoNet on 1-shot CUB and EuroSAT,
but reduces 1-shot ISIC accuracy; ProtoNet has the highest mean in all 5-shot
targets. Additional query context provides no reliable accuracy gain.
Rescue/break and boundary analyses characterize where predictions change,
while corruption, source-shift and scorer studies test the limits of the
intervention. These controlled comparisons, the query-context cost analysis,
and the replicated account of decision changes constitute the main empirical
contributions. Detailed protocols and secondary results are retained in the
appendices.

\section{Related work}
\label{sec:related}
Matching Networks, ProtoNet and Relation Networks classify from a support set
using fixed or learned similarity rules; MAML learns an initialization for
rapid parameter adaptation
\cite{vinyals2016matching,snell2017prototypical,sung2018relation,finn2017maml}.
FEAT adapts support-derived class representations, while PrototypeFormer uses
prototype tokens and within-class supports \cite{feat2020,prototypeformer}.
WIPT instead includes the current query in the same global-embedding context.

Single-query conditioning is not new: CrossTransformers and QPN use spatial
support--query matching, and HCPNet incorporates query information into
prototype formation \cite{crosstransformer2020,qpn2021,hcpnet2023}.
FSL-PRS, PRSN, RDProtoFusion, APPL and EfficientFSL use query sets alongside
mechanisms such as rectification, self-training or representation adaptation
\cite{fslprs2024,prsn2025,rdprotofusion2024,appl2024,efficientfsl2026}.
Our narrower setting excludes these additional mechanisms to study joint
query participation under a shared frozen representation. The support-only
control is not a reproduction of FEAT or PrototypeFormer.

FWT and StyleAdv motivate source-domain variation, while task-specific
adapters offer another route to target adaptation
\cite{tseng2020fwt,styleadv2023,taskadapters2022}. We test one source-only
appearance curriculum while freezing the encoder. \ref{app:related}
provides the full comparison, including Table~\ref{tab:positioning}.

\section{Method}
\label{sec:method}

\subsection{Problem setting and test-time prototype adaptation}
An $N$-way $K$-shot episode contains a support set
$S=\{(x_i,y_i)\}_{i=1}^{NK}$ with $K$ labelled examples per class and a set of
unlabelled query images $Q$. Let $f_\theta(\cdot)$ denote the image encoder and
let every image be represented by one global embedding in $\mathbb{R}^{D}$.
The encoder parameters $\theta$ are frozen throughout all controlled
experiments. Target support labels are used in the standard few-shot manner to
form class representatives; target query labels are never used by a classifier.
The task is closed-set: every query belongs to one of the support classes.

We distinguish \emph{prototype adaptation} from parameter adaptation. ProtoNet
uses one fixed support-derived class mean for every query. WIPT instead changes
the episode-specific class representative as a function of the current query,
but performs no target-time gradient update, backbone tuning, or pseudo-label
self-training.

\subsection{Shared frozen encoder}
All controlled methods use ViT-Small/16 with embedding dimension $D=384$. The
encoder is the \texttt{timm} checkpoint
\path{vit_small_patch16_224.augreg_in21k_ft_in1k}, pretrained on
ImageNet-21k and fine-tuned on ImageNet-1k. Its parameters have
\texttt{requires\_grad=False} throughout head training and evaluation. This is
central to the comparison: ProtoNet, the support-only Transformer and WIPT see
the same image representation, so differences arise only after the frozen
embedding has been produced. The implementation constructs the encoder with
\texttt{num\_classes=0}, using its default global feature output without an
additional projection before the episodic head.

\subsection{Single-query WIPT}
For a query embedding $\mathbf{q}\in\mathbb{R}^{D}$ and support embeddings
$\mathbf{s}_1,\ldots,\mathbf{s}_{NK}\in\mathbb{R}^{D}$, WIPT forms
\begin{equation}
\mathbf{T}^{(0)} =
\big[\,\mathbf{q};\ \mathbf{s}_1;\ \ldots;\ \mathbf{s}_{NK}\,\big]
\in \mathbb{R}^{(1+NK)\times D}.
\end{equation}
The sequence is passed through $L$ pre-norm Transformer layers,
\begin{equation}
\mathbf{T}^{(\ell)} =
\mathrm{Layer}_{\ell}\!\left(\mathbf{T}^{(\ell-1)}\right),
\qquad \ell=1,\ldots,L,
\end{equation}
followed by layer normalisation. Self-attention therefore allows both directions
of interaction: the transformed supports depend on the query and the
transformed query depends on the supports. We denote the outputs by
$\tilde{\mathbf{s}}_i$ and $\tilde{\mathbf{q}}$.

A class prototype is the mean of the transformed support embeddings,
\begin{equation}
\mathbf{p}_c(\mathbf{q})=
\frac{1}{K}\sum_{i:y_i=c}\tilde{\mathbf{s}}_i,
\end{equation}
and classification uses negative Euclidean distance,
\begin{equation}
\mathrm{score}(c)=-\left\lVert
\tilde{\mathbf{q}}-\mathbf{p}_c(\mathbf{q})
\right\rVert_2.
\end{equation}
The trainable episodic head is optimized with cross-entropy. The standard model
uses $L=2$ layers and six attention heads. Each pre-norm block has residual
self-attention and a residual feed-forward network of widths
$384\to1536\to384$, with GELU and dropout $0.1$. A final layer normalization
follows the two blocks. No positional or query/support type embeddings are
added. Source-query cross-entropy is averaged over all 75 queries in an episode;
one optimizer update is made per episode. We use the notation
\textbf{WIPT-2} here for this two-layer reference when discussing depth
ablations. The closest conceptual distinction from spatial single-query methods
such as Cross\-Transformers and QPN is that WIPT performs the entire interaction
on one global embedding per image, enabling a direct frozen-representation
control.

\begin{figure}[!htb]
\centering
\includegraphics[width=\textwidth]{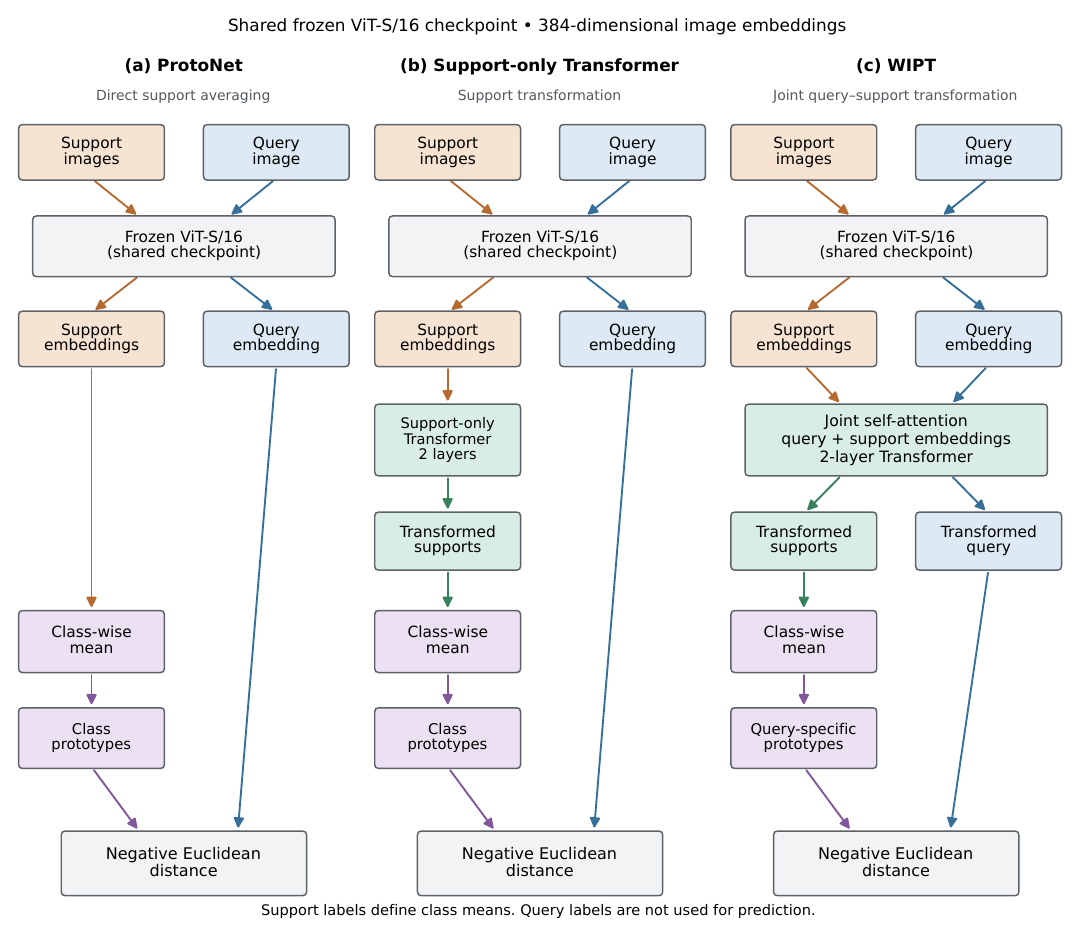}
\caption{Controlled prototype construction under one shared frozen ViT-S/16
encoder. ProtoNet averages support embeddings directly. The support-only
Transformer applies the same two-layer Transformer family to the support set
without exposing it to the query. WIPT jointly transforms one query and the
supports, then averages transformed supports class-wise to obtain query-specific
prototypes. Arrows distinguish the support and query pathways; class-wise
averaging follows support transformation in both Transformer methods.
Classification uses negative Euclidean distance. Only the episodic
Transformer heads are trained.}
\label{fig:arch}
\end{figure}

\subsection{Matched support-only control}
The support-only Transformer uses the same two-layer, six-head Transformer
configuration and dropout $0.1$ as WIPT, but processes only the support
embeddings. Its transformed class means are therefore episode-specific but not
query-specific, and the raw frozen query embedding is classified against those
means. This is the principal capacity-matched control: WIPT versus support-only
measures the effect of joint query participation under matched head capacity.
It changes both the support transformation and the query representation used
for scoring, so it does not isolate prototype conditioning alone. WIPT versus
ProtoNet compares the complete joint transformation with no learned episodic
transformation. The heads are trained separately; identical architecture does
not mean identical learned weights.

\subsection{Query context and computational cost}
\label{sec:multiquery_method}
The default WIPT processes one query with all $NK$ supports. For the context
comparison, we jointly process $g$ queries with the same supports, producing
group-conditioned prototypes. Queries are shuffled before grouping, with no
query-label access. Group-specific heads use five training seeds:
$g\in\{1,2,3,4,5\}$ for 5-shot and an exploratory $g\in\{1,3,5\}$ follow-up
for 1-shot. The $g=1$ path is the standard WIPT computation.

All groups use the same 3,549,696-parameter head. We measure analytical
attention-token pairs, head-only latency and peak CUDA allocated memory;
these exclude image encoding and data loading. \ref{app:query}
details grouping, training and the cost calculation, and \ref{app:cost}
reports the full benchmark. Lower attention-pair counts do not imply
proportional latency reductions.

\section{Experimental design}
\label{sec:setup}

\subsection{Datasets and cross-domain protocol}
We episodically train the trainable heads on miniImageNet using a
project-specific deterministic partition of its 100 classes. The WordNet
synset identifiers are sorted: the first 64 classes form the training set,
the next 16 the validation set and the final 20 the test set. Each class
contains 600 images, giving 38,400 training, 9,600 validation and 12,000 test
images. This partition differs from the usual Ravi--Larochelle class
assignment. Held-out miniImageNet serves as an in-domain reference.

Cross-domain evaluation uses Kaggle-distributed copies of CUB-200-2011
(200 classes), EuroSAT (10 classes, 27,000 images) and ISIC 2019
(8 classes, 25,331 images), representing fine-grained natural images,
satellite imagery and dermoscopic imagery, respectively. We sample target
episodes directly from the class folders, without a conventional target-domain
training/validation/test partition or target-domain parameter updates.
All images are converted to RGB. Each episode samples five classes and
$K+15$ distinct image indices per class without replacement: $K$ images form
the support set and 15 form the query set. Support and query indices are
therefore disjoint within an episode. Only support labels are supplied to
the classifier; unlabelled queries are processed individually or in groups
according to the experimental condition.
Because miniImageNet is derived from ImageNet and the encoder uses external
ImageNet pretraining, classes held out from episodic training are not
guaranteed to be unseen during encoder pretraining. The controlled comparisons
share this representation and do not establish performance under
pretraining-disjoint transfer.

\subsection{Controlled baselines and exact pipeline}
ProtoNet is the main fixed-prototype baseline. It has no trainable episodic
parameters. For evaluation, the exact encoder state stored in the WIPT
checkpoint is loaded into the ViT architecture, verified and frozen; ProtoNet
then forms class means directly from those shared embeddings. The support-only
Transformer and WIPT train only their episodic heads. The primary comparison fixes
Euclidean class-mean scoring across ProtoNet, support-only and WIPT, allowing
joint query participation to be assessed under one representation and scoring
rule. A separate five-training-seed factorial sensitivity study, reported in
\ref{app:factorial}, crosses raw, support-only and joint/WIPT
adaptation with Euclidean, cosine and learned relation scoring. These global
embedding controls are not full reproductions of Matching Networks, Relation
Networks, FEAT or PrototypeFormer.

Algorithm~\ref{alg:pipeline} in \ref{app:pipeline} gives the complete training and evaluation sequence.

\FloatBarrier
\subsection{Training and statistical evaluation}
Validation and evaluation images are resized to 256 pixels, centre-cropped to
$224\times224$, and normalized by the ImageNet mean
$(0.485,0.456,0.406)$ and standard deviation $(0.229,0.224,0.225)$. Episodic
training uses random resized cropping to $224\times224$, random horizontal
flipping and colour jitter of $0.4$ for brightness, contrast and saturation
before the same normalization.

All tasks are 5-way with 15 queries per class. We train matched 1-shot and
5-shot WIPT/support-only heads for at most 50 epochs, using 200 training and 50
validation episodes per epoch. Adam uses learning rate $10^{-4}$, zero weight
decay and cosine annealing with $T_{\max}=50$ and minimum learning rate $10^{-6}$. WIPT-2 uses six attention heads
and dropout $0.1$. The checkpoint attaining the highest validation accuracy
is retained. Independent models use training seeds $0,1,2,3,4$.

The principal cross-domain evaluation uses episode-sampling seeds
$42,123,2024,7,999$, with 300 episodes
per seed and domain, so each model sees 1,500 matched episodes per target. For
trainable models we report the mean and sample standard deviation across five
independent training runs. For paired differences, we report a two-sided 95\%
Student's $t$ interval across the five training-seed differences ($4$ degrees of
freedom). Episode-level confidence intervals are descriptive only and are not
substitutes for independent retraining. The same five-seed convention is used
for the 5-shot adaptation--scorer sensitivity study. For differences $d_r$, the
interval is $\bar d\pm t_{0.975,4}s_d/\sqrt{5}$. These intervals are conditional
on the fixed evaluation episode bank and quantify variation across trained
heads; they do not include uncertainty from drawing a new target benchmark.
They are unadjusted for the multiple domain, shot and comparator contrasts.
An interval containing zero indicates insufficient evidence of a directional
effect, not equivalence. Even an interval excluding zero should be interpreted
with its effect size and the small number of training seeds.

\subsection{Decision-change analysis and secondary experiments}
\label{sec:mechanism_protocol}
On the matched episodes, we count rescues (ProtoNet wrong, WIPT correct) and
breaks (ProtoNet correct, WIPT wrong). For ProtoNet accuracy $a$, rescue rate
$r$ among its errors and break rate $b$ among its successes,
\begin{equation}
\Delta a=(1-a)r-ab.
\label{eq:rescue_break}
\end{equation}
We locate these transitions relative to the true-class ProtoNet margin,
using separate equal-count bins for errors and successes. Query labels enter
only the post-prediction analysis. \ref{app:boundary} defines the
margin, binning and support-uncertainty diagnostics; these describe patterns
rather than establish causal pathways.

Secondary studies test controlled EuroSAT corruption
(\ref{app:corruption}), source-only shared-shift training (RQ4;
\ref{app:shift}), and a five-seed 5-shot adaptation--scorer grid
(\ref{app:factorial}). Selected-checkpoint ablations and diagnostics
remain explicitly exploratory.

\section{Results}
\label{sec:results}

\subsection{In-domain reference}
Held-out miniImageNet is close to saturation for the originally selected
5-shot checkpoints: support-only reaches $98.37\%$, frozen ProtoNet $98.34\%$
and WIPT $98.33\%$; the cosine-matching control reaches $98.47\%$. We treat
this only as an in-domain reference because the research questions concern the
non-saturated cross-domain targets.

\subsection{RQ1: replicated cross-domain performance}
\label{sec:crossdomain}
\begin{figure*}[!htb]
\centering
\includegraphics[width=\textwidth]{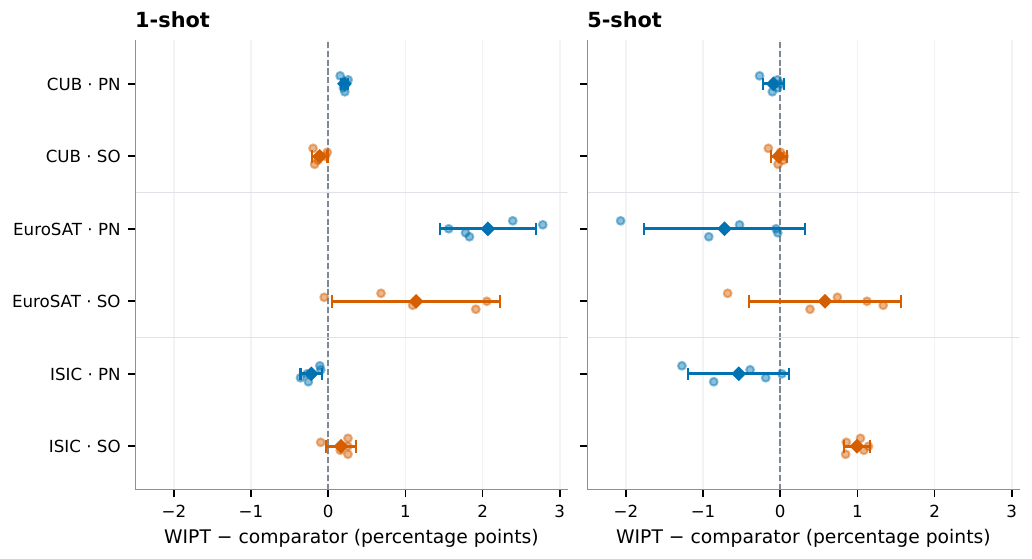}
\caption{Paired cross-domain accuracy effects in 1-shot and 5-shot evaluation.
PN denotes frozen ProtoNet and SO the support-only Transformer. Small points
show each of the five training-seed differences; diamonds and horizontal bars
show their mean and unadjusted 95\% Student's $t$ interval. The zero line marks
no accuracy difference. Absolute accuracies are reported in
Table~\ref{tab:crossdomain}.}

\label{fig:performance}
\end{figure*}

Table~\ref{tab:crossdomain} and Figure~\ref{fig:performance} separate
model-training variation from episode-sampling variation. ProtoNet is a fixed
point because it has no trainable episodic parameters; WIPT and support-only
statistics are across five independently trained heads.

In 5-shot evaluation, frozen ProtoNet has the highest mean accuracy on all three
target domains. WIPT--ProtoNet differences are $-0.089$ points on CUB,
$-0.721$ on EuroSAT and $-0.538$ on ISIC; none of the corresponding 95\%
training-seed intervals excludes zero. The capacity-matched comparison is more
specific. WIPT differs little from support-only on CUB, positive but
variable on EuroSAT ($+0.581$ points; four of five runs positive), and
consistently higher on ISIC ($+0.994$ points; five of five runs positive). The
5-shot ISIC WIPT--support-only interval is the only 5-shot interval that excludes
zero, providing the clearest replicated evidence that query participation adds
value beyond the same support transformation.

The 1-shot setting changes the ordering. WIPT is above ProtoNet on CUB by
$+0.210$ points and on EuroSAT by $+2.068$ points, with all five training runs
positive in both domains; it is lower on ISIC by $-0.218$ points, again in all
five runs. All three 1-shot WIPT--ProtoNet intervals exclude zero. Against the
support-only control, WIPT is lower on CUB and higher on EuroSAT, with both
unadjusted intervals excluding zero; the ISIC interval includes zero. The
EuroSAT WIPT--support-only difference is $+1.139$ points with a 95\% interval
of $[+0.052,+2.226]$ and four of five runs positive, so this is a less stable
contrast than the 1-shot gain over ProtoNet. Query conditioning is therefore neither uniformly
beneficial nor uniformly harmful; its value depends jointly on target domain
and shot count.

\begin{table*}[!htb]
\centering
\footnotesize
\setlength{\tabcolsep}{4pt}
\caption{Replicated cross-domain accuracy (\%) for the key controlled methods.
ProtoNet (PN) is a fixed point estimate because it has no trainable episodic
parameters. Support-only (SO) and WIPT are mean $\pm$ sample standard deviation
over five independent training seeds. Difference columns are paired
training-seed mean $\pm$ standard deviation. Boldface marks the highest mean
accuracy within each shot/domain row. $\dagger$ indicates that the corresponding
unadjusted 95\% training-seed interval for the difference excludes zero;
no multiple-comparison correction is applied.}
\label{tab:crossdomain}
\begin{tabular}{c l c c c c c}
\toprule
Shot & Domain & PN & SO & WIPT & $\Delta$ WIPT--PN & $\Delta$ WIPT--SO \\
\midrule
5 & CUB & \textbf{97.88} & $97.81\pm0.03$ & $97.79\pm0.11$ & $-0.09\pm0.11$ & $-0.02\pm0.08$ \\
5 & EuroSAT & \textbf{82.76} & $81.46\pm0.08$ & $82.04\pm0.84$ & $-0.72\pm0.84$ & $+0.58\pm0.79$ \\
5 & ISIC & \textbf{38.92} & $37.39\pm0.49$ & $38.38\pm0.53$ & $-0.54\pm0.53$ & $+0.99\pm0.13^{\dagger}$ \\
\addlinespace
1 & CUB & $93.26$ & $\boldsymbol{93.58\pm0.05}$ & $93.47\pm0.04$ & $+0.21\pm0.04^{\dagger}$ & $-0.11\pm0.08^{\dagger}$ \\
1 & EuroSAT & $62.09$ & $63.02\pm1.35$ & $\boldsymbol{64.16\pm0.50}$ & $+2.07\pm0.50^{\dagger}$ & $+1.14\pm0.88^{\dagger}$ \\
1 & ISIC & \textbf{28.40} & $28.01\pm0.23$ & $28.18\pm0.11$ & $-0.22\pm0.11^{\dagger}$ & $+0.17\pm0.15$ \\
\bottomrule
\end{tabular}
\end{table*}

\FloatBarrier
\subsection{RQ2: additional query context and cost}
\label{sec:multiquery_results}
\begin{figure*}[!htb]
\centering
\includegraphics[width=\textwidth]{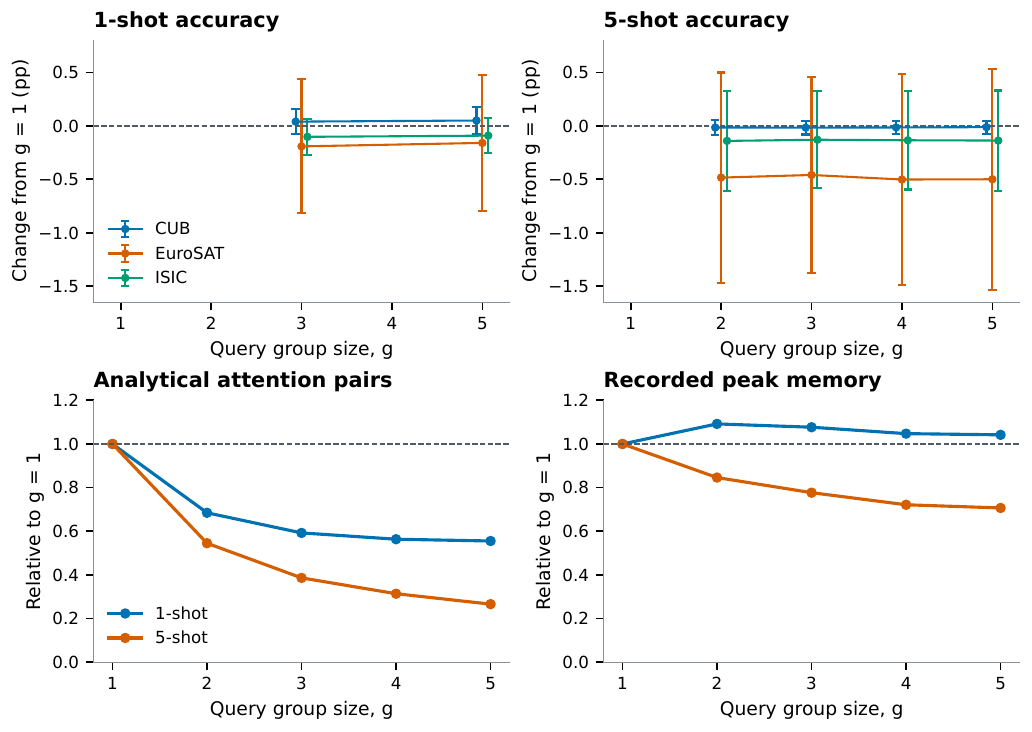}
\caption{Multi-query accuracy and compute trade-offs. Top: accuracy change
relative to $g=1$ in the 1-shot sparse follow-up and full 5-shot sweep, with
unadjusted 95\% training-seed intervals. Bottom: analytical token-pair count and
recorded peak memory, each normalized to its shot-specific $g=1$ benchmark.
Timing benchmarks exist for $g=2,4$ in 1-shot even though those groups were not
included in the 1-shot accuracy study. Lower token-pair counts need not imply
lower latency; full timings appear in \ref{app:cost}.}

\label{fig:multiquery}
\end{figure*}

Figure~\ref{fig:multiquery} summarizes the context comparison. All paired
95\% training-seed intervals include zero: mean 5-shot changes relative to
$g=1$ range from $-0.012$ to $-0.017$ points on CUB, $-0.459$ to $-0.502$
on EuroSAT, and $-0.130$ to $-0.141$ on ISIC. The 1-shot follow-up also
shows no reliable gain. This does not establish equivalence: the 5-shot
EuroSAT $g=5$ interval is $[-1.536,+0.536]$ points.

In 5-shot, $g=5$ reduces attention-token pairs from 101,400 to 27,000
(73\%) and recorded peak allocated memory from 75.79 to 53.58 MiB (29\%).
Latency is non-monotonic: forward medians are 2.40 ms at $g=1$ and 3.09 ms
at $g=5$. Memory is roughly flat or slightly higher with grouping in 1-shot.
Thus $g=1$ supports independent, streaming-compatible predictions, while
grouping can amortize support computation without a demonstrated accuracy
benefit. Full accuracy and cost results appear in
\ref{app:query} and~\ref{app:cost}.

\FloatBarrier

\subsection{RQ3: shot, domain and decision changes}
\label{sec:mechanism_results}
\begin{figure*}[!htb]
\centering
\includegraphics[width=\textwidth]{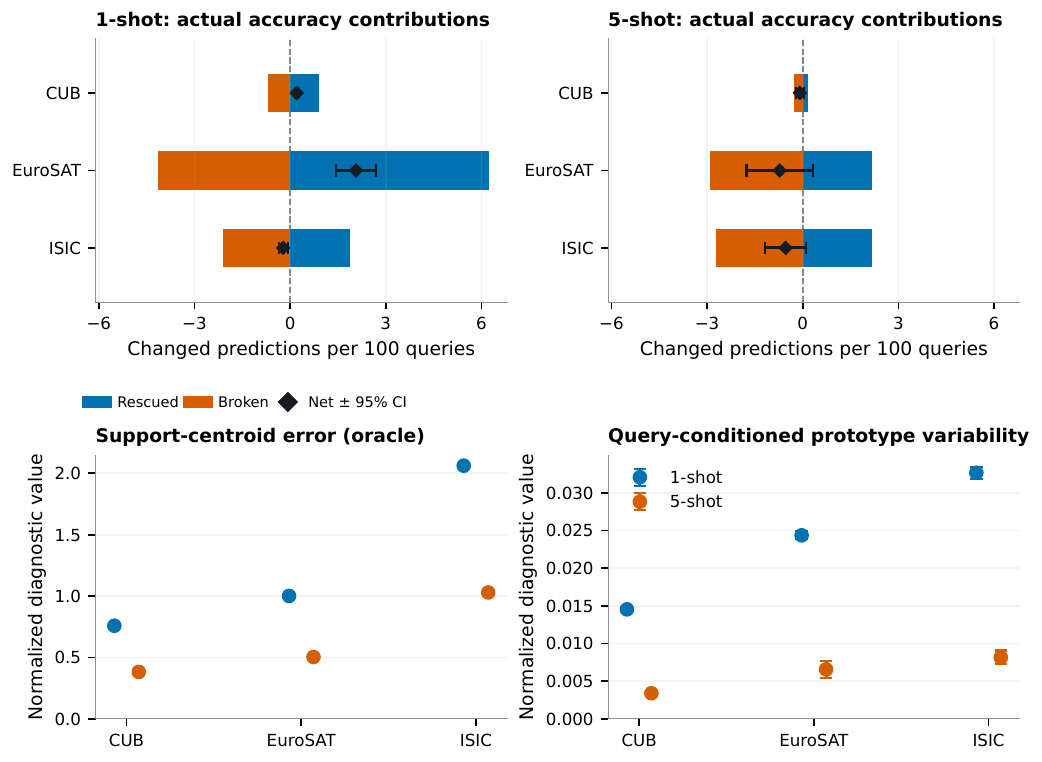}
\caption{Decision-change decomposition and support diagnostics. Top: rescued
and broken predictions per 100 queries; black diamonds show the net accuracy
effect, with a 95\% interval across five WIPT training seeds. The two bars have
the same denominator, so their signed sum is the net effect. Bottom left:
analysis-only support-to-labelled-query centroid error. This fixed-encoder
quantity is identical across WIPT seeds and has no training-seed error bar.
Bottom right: query-conditioned prototype variability with training-seed
intervals. Both diagnostics use inter-class query-centroid distance for
normalization; query labels are used only after prediction.}
\label{fig:shotmechanism}
\end{figure*}

Support-to-query centroid error approximately halves from 1-shot to 5-shot,
and query-conditioned prototype variability also decreases
(Figure~\ref{fig:shotmechanism}). These are descriptive associations with
support size, not causal explanations of accuracy.

The rescue/break decomposition accounts for the net effects. Per 100 queries,
1-shot EuroSAT has 6.223 rescues and 4.155 breaks, giving $+2.068$ points.
In 5-shot, 2.183 rescues and 2.905 breaks give $-0.721$ points; the baseline
error pool falls from 37.91\% to 17.24\%. Five-shot CUB is near saturation:
0.178 rescues are outweighed by 0.266 breaks. On 1-shot ISIC, 1.879 rescues
are outweighed by 2.098 breaks. These comparisons use a common denominator;
conditional rescue and break rates cannot simply be subtracted.

Correctness transitions concentrate near the ProtoNet boundary. The 20\%
of queries with smallest absolute margin contain 100.0\%, 68.4\% and 96.5\%
of all rescues and breaks on 1-shot CUB, EuroSAT and ISIC; the 5-shot values
are 100.0\%, 91.1\% and 91.4\%. The separate error/success curves in
Figure~\ref{fig:boundary_all} (\ref{app:boundary}) show higher
transition probabilities near zero. The sign of a transition alone is
structurally constrained by baseline correctness and is not mechanistic evidence.
Episode-level geometry correlations with gain are weak (maximum absolute
coefficient 0.146); neither these associations nor boundary localization
establish why an individual transformation helps or harms.

\FloatBarrier

\subsection{Controlled corruption response}
\label{sec:corruption_results}
The five-seed EuroSAT stress test shows conditional robustness. Combined
brightness, contrast and noise changes the mean WIPT--ProtoNet effect from
$-0.26$ points at severity 3 to $+0.75$ and $+1.02$ at severities 4 and 5,
with all five runs positive at the latter severities. Brightness/contrast
alone mostly favours ProtoNet or gives a tie; Gaussian-noise effects remain
small. WIPT is above the matched support-only control on average in every
condition. Thus robustness relative to transformed supports does not imply
uniform superiority over raw class means. \ref{app:corruption}
retains the schedule, full interpretation and Figure~\ref{fig:corruption}.

\subsection{RQ4: source-only shared-shift training}
\label{sec:shift_results}
A shared pseudo-domain appearance curriculum gives no evidence of improvement
over standard WIPT: all six domain/shot paired intervals include zero.
Mean changes range from $-0.399$ to $-0.055$ points in 1-shot and from
$-0.419$ to $+0.018$ in 5-shot. Query participation can still help within
the shift-trained comparison: WIPT exceeds shift-trained support-only by
$+0.82$ points on 5-shot EuroSAT and $+1.47$ on ISIC, with all five runs
positive. The negative result concerns improvement over ordinary WIPT.
\ref{app:shift} gives the full protocol and Table~\ref{tab:shift};
the single-seed mixed-shift screen remains exploratory
(\ref{app:mixedshift}).

\subsection{Sensitivity to adaptation and scoring}
\label{sec:factorial_results}
The five-seed $3\times3$ grid identifies no stronger absolute alternative
to raw Euclidean scoring. Cosine-trained WIPT is below raw and support-only
cosine in all runs on every target; its EuroSAT difference from raw cosine
is $-25.317\pm4.818$ points (95\% interval). WIPT with relation scoring
improves support-only relation on CUB by $+8.515\pm2.382$ points, but
remains well below Euclidean accuracy. These are complete scorer/training
configurations, not isolated metric substitutions. \ref{app:factorial}
retains the protocols, all condition means and paired contrasts, including
the distinction from inference-only cosine ablations.

\section{Discussion}
\label{sec:discussion}

\subsection{A conditional benefit, not a universal replacement}
The central result is not that attention should replace class averaging. Frozen
ProtoNet remains a remarkably strong 5-shot baseline, and WIPT does not improve
it on average in any of the three 5-shot targets. The more informative
comparison is conditional. When support is scarce, WIPT improves ProtoNet on
CUB and EuroSAT in every independent 1-shot run, whereas the same intervention
is consistently harmful on ISIC. Against the capacity-matched support-only
Transformer, the clearest 5-shot gain appears on ISIC. These contrasts quantify the effect of joint query participation while
showing that its value depends on the target domain and amount of labelled
support. They do not separate query-side adaptation from support-side conditioning.

This framing also clarifies the contribution relative to prior query-aware
methods. WIPT is intentionally smaller in scope than systems that use spatial
matching, transductive query sets, pseudo-label rectification or target-time
feature updates. Its purpose is to expose what one query can contribute when
those additional mechanisms are absent. The result is useful precisely because
it includes regimes in which the intervention helps, does little, or hurts.

\subsection{No detected accuracy benefit from additional query context}
The multi-query experiment tests whether WIPT benefits from query-set context
rather than assuming that a single query is intrinsically sufficient. Across the
tested range, adding unlabelled queries to the WIPT sequence does not produce a
reliable cross-domain accuracy gain in either shot regime. Larger groups reduce
analytical attention work per query by amortizing the support tokens, but the
measured head-only latency is non-monotonic. The practical case for $g=1$ is
therefore its support for independent, streaming-compatible inference.
The study detects no reliable gain from the tested $g>1$ variants, but it
does not show that their accuracies are equivalent. This does not
contradict query-set methods that use additional context through pseudo-labels,
rectification or target adaptation; those methods operate in a different
inference regime.

\subsection{Localization and the balance of decision changes}
The query-level analysis shows where the intervention matters. WIPT changes
uncertain ProtoNet decisions much more frequently than confident ones. The
separate rescue and break curves show this descriptive localization;
their signs are not offered as independent evidence of a mechanism.
Equation~\ref{eq:rescue_break} then accounts exactly for the aggregate
accuracy difference by weighting each transition rate by its opportunity pool.

Additional support examples coincide with lower centroid error and less
query-conditioned prototype variability. They also reduce the baseline error
pool, leaving fewer opportunities for rescue. This provides a coherent
description of the shot-dependent results, although a causal test would need
to vary the query and support adaptation pathways separately. ISIC further
shows that a large conditioning response is not intrinsically helpful: its
large support uncertainty and prototype variability coexist with net accuracy
losses relative to ProtoNet. The selected-checkpoint normalized-margin
diagnostic in \ref{app:exploratory} likewise gives no evidence of a
uniform margin improvement.

\subsection{What the negative extensions tell us}
The corruption and source-shift experiments place useful limits on the method.
Under controlled EuroSAT corruption, WIPT is often more robust than the matched
support-only Transformer, but only the severe combined-corruption regime crosses
reliably above ProtoNet. Likewise, the source-only shared pseudo-domain curriculum provides no
evidence of improvement over ordinary WIPT on the natural target datasets.
Merely exposing the episodic head to broader appearance variation is therefore
insufficient, in this setup, to establish a generally better cross-domain
adaptation rule under a fixed backbone.
Methods such as FWT and StyleAdv suggest that richer feature-level or
adversarial style variation may be more effective
\cite{tseng2020fwt,styleadv2023}.

The replicated $3\times3$ scorer study reaches a related conclusion from a
different direction. Query participation is not invariant to the downstream
metric. Training WIPT with the tested cosine configuration performs poorly on all
three domains, whereas
relation scoring produces a large gain over its support-only counterpart on CUB
but remains far below Euclidean accuracy. Euclidean class-mean scoring therefore
remains the strongest overall formulation, while the grid warns against treating
query conditioning as a plug-in improvement independent of the classifier that
uses the adapted prototypes. This is not a universal failure of cosine
similarity: temperature, normalization and scorer-specific tuning can change
its training behaviour, and an inference-only cosine substitution tests a
different question. Here, the factorial cosine condition also averages
individual support similarities rather than comparing normalized class means;
its poor accuracy should not be generalized to all cosine-prototype methods.

\subsection{Limitations and future work}
The study uses one source dataset, three target domains and one frozen ViT-S/16
backbone. It is designed to characterize a controlled intervention, not to provide a leaderboard
comparison with systems that fine-tune larger encoders, use spatial features or
adapt transductively on target queries. The multi-query study considers groups
of at most five queries and uses a sparse $g=1,3,5$ confirmation in 1-shot. The
$3\times3$ adaptation--scorer grid is fully replicated but only in 5-shot. The
shared-shift experiment tests one feasible source-only curriculum, while the
mixed-shift follow-up has only one training seed and remains exploratory.

The corruption family is also intentionally narrow, and the selected-checkpoint
geometry, support-noise and design-ablation analyses should not be read as
between-training-run evidence. Although the boundary analysis covers all
three targets in both shot regimes, it still reflects one backbone and one
source training distribution. Future work should repeat the controlled study
with additional source datasets, frozen and fine-tuned encoders, larger query
sets and spatial query-aware methods under the same episodic protocol. A useful
next step would be to test whether the boundary-local rescue/break pattern
persists when the representation itself is allowed to adapt. Another
priority is a control that transforms the query without allowing it to alter
support prototypes, paired with a control that allows prototype conditioning
while holding the scoring query fixed. This would separate the two effects
combined in WIPT.

\section{Conclusion}
\label{sec:conclusion}
WIPT implements a simple form of test-time prototype adaptation: one target query
and the labelled support set interact in global embedding space, while the image
encoder and model parameters remain fixed. Five independent training runs show
that this intervention is useful in selected regimes rather than uniformly
superior to ProtoNet. Its clearest ProtoNet gain occurs on 1-shot EuroSAT, and
query participation also consistently improves the matched support-only
Transformer on 5-shot ISIC.

The broader experiments delimit that result. None of the tested $g>1$
variants provides a reliable accuracy gain over processing queries
independently, although the intervals do not establish equivalence. Across
CUB, EuroSAT and ISIC, the largest decision changes are concentrated near
ProtoNet's boundary, and the aggregate outcome depends on the balance between
rescued errors and broken successes. Source-only shift training provides no
evidence of improvement over standard WIPT, while the replicated scorer grid
shows that the effect is sensitive to the downstream metric. The contribution is
therefore not a universal accuracy claim, but a controlled account of when
query-conditioned prototypes help difficult cross-domain few-shot decisions,
when fixed prototypes remain preferable, and where the adaptation remains
fragile.

\FloatBarrier
\clearpage
\appendix
\renewcommand{\thefigure}{\Alph{section}.\arabic{figure}}
\renewcommand{\thetable}{\Alph{section}.\arabic{table}}
\renewcommand{\theHfigure}{\Alph{section}.\arabic{figure}}
\renewcommand{\theHtable}{\Alph{section}.\arabic{table}}
\setcounter{figure}{0}
\setcounter{table}{0}

\section{WIPT design ablations}
\label{app:ablations}
The two-layer standard formulation is denoted WIPT-2. These design ablations use
the originally selected 5-shot WIPT checkpoint and are exploratory architecture
checks, not five-training-seed comparisons. Trained variants use the same frozen
ViT-S/16 encoder and episodic training protocol as WIPT-2. Inference-only
variants reuse the WIPT-2 checkpoint, so their differences arise only from the
decision rule.

WIPT-4 and WIPT-6 change only the number of Transformer layers. Residual
refinement averages the transformed query with its original encoder embedding.
Combined refinement adds the residual path, input layer normalization, support
dropout $0.1$ and attention temperature 2.0. Query-weighted pooling replaces
the class-wise mean with a weighted mean derived from final-layer
query-to-support attention. Multi-prototype variants retain each transformed
support instance and score by the closest support. The margin-training variant
adds $0.5\,\mathrm{softplus}[-(s_y-\max_{c\ne y}s_c)]$ to cross-entropy.

\begin{table}[!htb]
\centering
\small
\caption{Complete cross-domain accuracy (\%) for WIPT design ablations. Values
are means over 1,500 matched evaluation episodes per target; WIPT-2 is the
reference.}
\label{tab:ablation_full}
\begin{tabular}{lccc}
\toprule
Variant & CUB & EuroSAT & ISIC \\
\midrule
WIPT-2 & 97.864 & 82.324 & 38.753 \\
\addlinespace[2pt]
\multicolumn{4}{l}{\emph{Trained architecture/objective variants}} \\
WIPT-4 & 97.610 & 79.203 & 37.561 \\
WIPT-6 & 97.815 & 81.756 & 38.637 \\
Residual refinement & 97.805 & 80.401 & 37.240 \\
Combined refinement & 97.845 & 82.185 & 38.419 \\
Margin-training objective & 97.820 & 82.256 & 38.239 \\
\addlinespace[2pt]
\multicolumn{4}{l}{\emph{Inference-only variants using WIPT-2 checkpoint}} \\
Query-weighted pooling, $T=1.0$ & 97.862 & 82.333 & 38.728 \\
Query-weighted pooling, $T=0.5$ & 97.860 & 82.343 & 38.719 \\
Query-weighted pooling, $T=0.25$ & 97.863 & 82.350 & 38.692 \\
Cosine scoring & 97.959 & 81.963 & 38.564 \\
Multi-prototype, Euclidean & 97.325 & 77.540 & 35.788 \\
Multi-prototype, cosine & 97.322 & 77.536 & 35.791 \\
\bottomrule
\end{tabular}
\end{table}

\FloatBarrier
No modification improves all three target domains. Increasing depth to four
layers reduces EuroSAT by 3.12 points, six layers remain 0.57 points below
WIPT-2, query-weighted pooling is almost neutral, inference-only cosine scoring trades a small
CUB gain for losses elsewhere, and multi-prototype scoring is consistently
weaker. We therefore retain the two-layer, class-mean, Euclidean model.

The inference-only cosine result on EuroSAT (81.963\%) uses the original
Euclidean-trained WIPT-2 checkpoint. It should not be equated with the
53.425\% mean for five separately cosine-trained heads in
\ref{app:factorial}. The factorial condition additionally uses class-wise averaging of
individual support cosine similarities, whereas this ablation normalizes the
class-mean prototype. These results differ in both training and aggregation
and should not be interpreted as inconsistent evaluations of one scorer.

\FloatBarrier

\section{Five-seed adaptation--scorer factorial grid}
\setcounter{figure}{0}
\setcounter{table}{0}
\label{app:factorial}

\label{sec:factorial_protocol}
The primary RQ1 comparison fixes Euclidean class-mean scoring to assess joint query
participation. As a secondary sensitivity analysis, we test whether that
conclusion depends on the decision rule by crossing three adaptation paths
--- raw, support-only and joint/WIPT --- with Euclidean, cosine and a learned
relation scorer in the 5-shot regime. Raw Euclidean and raw cosine have no
trainable episodic parameters; raw relation trains only the relation scorer.
The support-only and joint conditions train the corresponding episodic head
under each scorer. All trainable factorial conditions use seeds
$0,1,2,3,4$, and evaluation uses the same 1,500 matched target episodes per
domain as the primary study. The Euclidean support-only and WIPT cells reuse
the standard five-seed checkpoints. We report training-seed means and 95\%
Student's $t$ intervals, paired WIPT contrasts and positive-run counts. Because
the grid is a secondary sensitivity analysis with multiple contrasts, we
emphasize effect sizes, intervals and replication direction rather than an
exhaustive hypothesis-testing screen. This grid trains each head with its
specified scorer. In contrast, the cosine ablation in \ref{app:ablations}
changes the scoring rule only at inference for one Euclidean-trained checkpoint.
They also use different cosine aggregation rules. In the factorial grid,
each query and each support vector are $\ell_2$-normalized and the cosine
similarities are averaged within each class:
\begin{equation}
s_c^{\mathrm{cos}}(\mathbf q)=\frac{1}{K}\sum_{i:y_i=c}
\frac{\mathbf q^\top\mathbf s_i}{\|\mathbf q\|_2\|\mathbf s_i\|_2},
\label{eq:cosine_matching}
\end{equation}
using the appropriate raw or transformed vectors. The logits have no learned
scale or additional temperature (equivalently $T=1$). In the inference-only
ablation, cosine similarity is instead computed between the transformed query
and the normalized class-mean prototype. Averaging normalized support
similarities is not generally the same operation as normalizing a class mean.

The relation scorer concatenates a query with a class-mean prototype and
applies a $768\to256\to128\to1$ multilayer perceptron with ReLU hidden
activations and a sigmoid output. It has 229,889 trainable parameters and is
trained with mean squared error against one-hot class targets. Euclidean and
cosine models use cross-entropy. The grid therefore compares complete
scorer/training configurations; it is not an isolated metric substitution.
\subsection{Full condition results}
To test whether the main conclusions depend on the downstream decision rule, we
evaluate a 5-shot $3\times3$ grid with adaptation in
\{raw, support-only, joint/WIPT\} and scoring in
\{Euclidean, cosine, learned relation\}. Raw Euclidean and raw cosine are fixed
rules with no trained episodic head. Raw relation trains the relation scorer,
and all Transformer-based rows are trainable. Every trainable condition is
replicated with independent training seeds $0$--$4$ and evaluated on the same
1,500 matched target episodes per domain. Table~\ref{tab:factorial} reports
condition means with 95\% Student's $t$ intervals across training seeds where
applicable; Table~\ref{tab:factorial_contrasts} reports paired WIPT contrasts
and the number of positive WIPT runs.

\begin{table*}[!htb]
\centering
\footnotesize
\setlength{\tabcolsep}{4pt}
\caption{Five-seed 5-shot $3\times3$ adaptation--scorer grid (accuracy, \%).
Trainable entries are mean $\pm$ 95\% Student's $t$ interval across independent
training seeds. Raw Euclidean and raw cosine are fixed rules and therefore have
no training-seed interval.}
\label{tab:factorial}
\begin{tabular}{l l c c c}
\toprule
Adaptation & Scorer & CUB & EuroSAT & ISIC \\
\midrule
Raw & Euclidean & \textbf{97.880} & \textbf{82.763} & \textbf{38.918} \\
Raw & Cosine & 97.468 & 78.742 & 36.239 \\
Raw & Relation & $79.058\pm0.879$ & $60.198\pm2.225$ & $27.983\pm1.623$ \\
\addlinespace
Support-only & Euclidean & $97.810\pm0.035$ & $81.461\pm0.104$ & $37.386\pm0.608$ \\
Support-only & Cosine & $97.385\pm0.019$ & $77.488\pm0.202$ & $35.927\pm0.107$ \\
Support-only & Relation & $72.594\pm1.743$ & $58.722\pm2.057$ & $26.505\pm0.864$ \\
\addlinespace
Joint/WIPT & Euclidean & $97.791\pm0.132$ & $82.041\pm1.041$ & $38.380\pm0.652$ \\
Joint/WIPT & Cosine & $95.350\pm0.556$ & $53.425\pm4.818$ & $33.019\pm0.597$ \\
Joint/WIPT & Relation & $81.109\pm1.497$ & $58.628\pm1.922$ & $27.406\pm1.085$ \\
\bottomrule
\end{tabular}
\end{table*}

\begin{table*}[!htb]
\centering
\footnotesize
\setlength{\tabcolsep}{4pt}
\caption{Paired WIPT contrasts in the five-seed factorial study. Values are
mean accuracy differences in percentage points $\pm$ 95\% Student's $t$
intervals across training seeds. The parenthesized count is the number of
training seeds for which WIPT has higher accuracy than the comparator.}
\label{tab:factorial_contrasts}
\begin{tabular}{l l c c}
\toprule
Domain & Scorer & WIPT $-$ Raw & WIPT $-$ Support-only \\
\midrule
CUB & Euclidean & $-0.089\pm0.132$ (0/5) & $-0.019\pm0.101$ (3/5) \\
CUB & Cosine & $-2.119\pm0.556$ (0/5) & $-2.035\pm0.565$ (0/5) \\
CUB & Relation & $+2.051\pm2.209$ (4/5) & $+8.515\pm2.382$ (5/5) \\
\addlinespace
EuroSAT & Euclidean & $-0.721\pm1.041$ (0/5) & $+0.581\pm0.985$ (4/5) \\
EuroSAT & Cosine & $-25.317\pm4.818$ (0/5) & $-24.063\pm4.698$ (0/5) \\
EuroSAT & Relation & $-1.571\pm3.713$ (1/5) & $-0.094\pm3.516$ (2/5) \\
\addlinespace
ISIC & Euclidean & $-0.538\pm0.652$ (1/5) & $+0.994\pm0.167$ (5/5) \\
ISIC & Cosine & $-3.220\pm0.597$ (0/5) & $-2.909\pm0.670$ (0/5) \\
ISIC & Relation & $-0.577\pm1.185$ (1/5) & $+0.901\pm1.045$ (4/5) \\
\bottomrule
\end{tabular}
\end{table*}

\FloatBarrier
Euclidean scoring remains the strongest absolute formulation in every target
domain. The five-seed contrasts also show that the effect of query
participation depends strongly on the scorer. With cosine scoring, WIPT is
lower than both raw and support-only variants in all five training runs on all
three domains; the EuroSAT drop is especially large. With relation scoring,
the clearest replicated benefit is on CUB: WIPT exceeds support-only relation by
$+8.515\pm2.382$ points with five of five runs positive. Its
$+2.051\pm2.209$ point mean advantage over raw relation is less stable
(four of five positive) and its interval includes zero. EuroSAT relation
contrasts are highly variable, while the ISIC WIPT--support relation contrast
is positive on average but also includes zero. The grid therefore shows
a scorer--adaptation interaction under the tested training settings without
identifying a stronger replacement
for the main Euclidean formulation.

\FloatBarrier
\section{Exploratory mixed pseudo-domain screening}
\setcounter{figure}{0}
\setcounter{table}{0}
\label{app:mixedshift}
The five-seed shared-shift experiment uses one pseudo-domain transform for both
support and query on shifted source episodes. As a low-cost follow-up, we also
screen a \emph{mixed} regime in which each shifted episode independently chooses
between the shared case and a cross case with separate support and query
pseudo-domains. As in the shared experiment, each training episode is clean
with probability 0.5 and shifted with probability 0.5, and shift strength
follows the same curriculum. This exploratory screen uses training seed 0;
Table~\ref{tab:mixedshift} compares it with the
matched ordinary WIPT seed 0 and is descriptive.

\begin{table}[!htb]
\centering
\small
\caption{Single-seed mixed pseudo-domain screen. $\Delta$ is mixed-shift WIPT
minus the matched ordinary WIPT seed-0 accuracy (percentage points).}
\label{tab:mixedshift}
\begin{tabular}{c l c c c}
\toprule
Shot & Domain & Standard seed 0 & Mixed seed 0 & $\Delta$ \\
\midrule
1 & CUB & 93.471 & 93.597 & +0.126 \\
1 & EuroSAT & 63.921 & 63.670 & -0.251 \\
1 & ISIC & 28.139 & 28.059 & -0.080 \\
\addlinespace
5 & CUB & 97.778 & 97.818 & +0.040 \\
5 & EuroSAT & 81.837 & 81.899 & +0.061 \\
5 & ISIC & 38.057 & 38.290 & +0.233 \\
\bottomrule
\end{tabular}
\end{table}

The mixed seed does not show a coherent large improvement: changes are small
and change sign across the 1-shot targets. The 5-shot changes are all positive
but at most $+0.233$ points and are based on one initialization. These exploratory results do not alter the five-seed RQ4 conclusion.

\FloatBarrier
\section{Exploratory selected-checkpoint diagnostics}
\setcounter{figure}{0}
\setcounter{table}{0}
\label{app:exploratory}

\subsection{Normalized prototype geometry}
For each episode, let $d_{\mathrm{correct}}$ be the query distance to its true
prototype and $d_{\mathrm{wrong}}$ the distance to the nearest incorrect
prototype. The query margin $d_{\mathrm{wrong}}-d_{\mathrm{correct}}$ is
normalized by mean inter-prototype distance. WIPT and ProtoNet have almost the
same mean normalized margin: $0.3238$ versus $0.3245$ on CUB, $0.1632$ versus
$0.1645$ on EuroSAT and $-0.0470$ versus $-0.0468$ on ISIC. Matched episode
margins are highly correlated ($r=0.998,0.987,0.994$). WIPT differs more from
the support-only control on CUB and EuroSAT, but does not systematically enlarge
the normalized margin beyond ProtoNet.

\begin{figure*}[!htb]
\centering
\includegraphics[width=\textwidth]{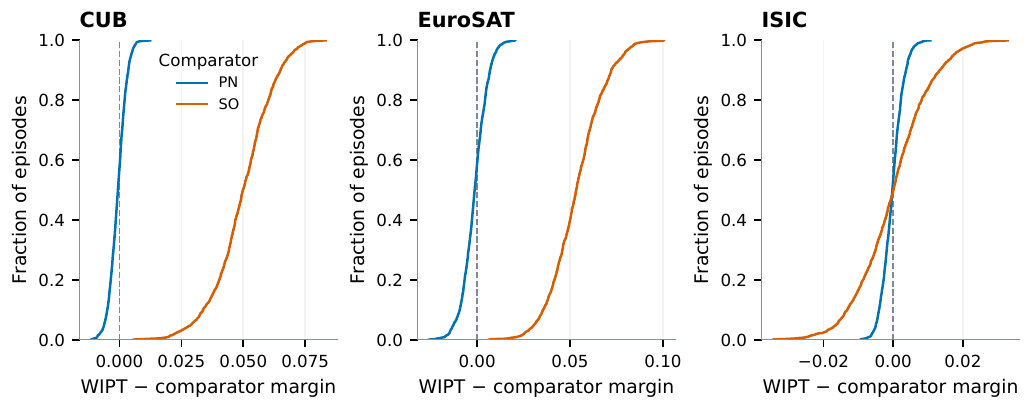}
\caption{Empirical distributions of paired episode-level normalized-margin
changes for the selected 5-shot checkpoint, relative to ProtoNet (PN) and
support-only (SO). Each curve uses 1,500 matched episodes. The horizontal
coordinate is a difference of normalized margins, not accuracy; no
training-seed uncertainty is inferred from these curves.}

\label{fig:geometry}
\end{figure*}

\FloatBarrier
\subsection{Support-noise pathway diagnostic}
In a separate selected-checkpoint 5-shot miniImageNet test diagnostic,
Gaussian noise is applied only to support images, while
query images remain clean. For WIPT, the transformed query is first computed
from the clean-support pass and then deliberately \emph{held fixed}; only the
prototype side is recomputed after replacing the clean support with noisy
support. This is a counterfactual pathway-isolation diagnostic, not the normal
WIPT inference procedure. It asks how much a support perturbation moves the
prototype when query-side movement has been removed from the comparison. The diagnostic uses 200 episodes for each of the
five evaluation seeds (1,000 episodes), with Gaussian standard deviations
0.18 and 0.38 at severities 3 and 5. Noise is added after reversing ImageNet
normalization; RGB values are clipped to $[0,1]$ and then renormalized.

The displacement ratio measures attenuation through support averaging. For
ProtoNet, its denominator is the mean displacement of raw frozen support
embeddings and its numerator is the mean prototype displacement. For WIPT,
both are measured \emph{after} joint transformation: the denominator averages
transformed-support displacements over queries, classes and shots, and the
numerator averages transformed-prototype displacements over queries and
classes. Each episode forms the ratio using a denominator stabilizer
$10^{-8}$; the reported value averages these episode ratios. Values are 0.662
versus 0.660 at severity 3 and 0.785 versus 0.807 at severity 5. Because the
WIPT denominator already includes the Transformer response, this ratio does
not measure amplification from raw frozen embeddings through the whole head. Their clean absolute
margins are 15.04 and 6.25, respectively, but these distances come from
different representation scales: WIPT includes a learned transformation and
final layer normalization. A smaller absolute margin therefore cannot, by
itself, establish a smaller robustness buffer.

\begin{figure*}[!htb]
\centering
\includegraphics[width=\textwidth]{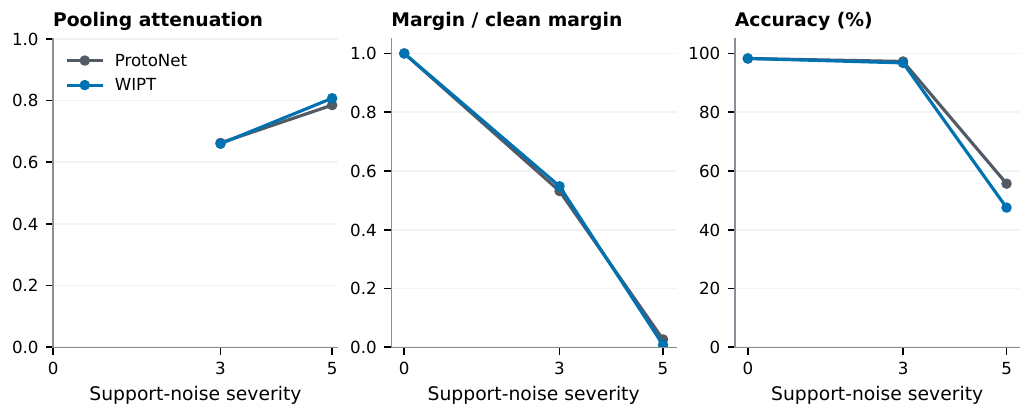}
\caption{Selected-checkpoint support-noise pathway diagnostic. The WIPT query
representation from the clean-support pass is held fixed while the noisy
support-derived prototypes are recomputed, isolating support-to-prototype
propagation. (a) Attenuation through class-wise support averaging in each method's
respective support representation. (b) Reported margin divided by the same model's clean margin, to avoid
comparing raw distance scales. (c) Accuracy. Points are selected-checkpoint
summaries; connecting lines are visual guides, and error bars are not shown.}
\label{fig:noise}
\end{figure*}

Figure~\ref{fig:noise} instead expresses each reported margin relative to that
model's clean value. At severity 5, the ratios are approximately 0.027 for
ProtoNet and 0.008 for WIPT; accuracies are 55.69\% and 47.53\%.
This is consistent with stronger relative margin collapse in this diagnostic,
but does not establish the cause of the accuracy gap or reproduce the full
noisy-query inference procedure. The diagnostic uses one fixed WIPT-2
reference checkpoint; its originating training seed cannot be established
because the original root-level checkpoint is no longer retained. The five
seeds here control evaluation episodes, not independent head training.
These selected-checkpoint summaries therefore do not quantify variation
across independently trained heads.

\FloatBarrier
\section{Additional computational-cost results}
\setcounter{figure}{0}
\setcounter{table}{0}
\label{app:cost}
Table~\ref{tab:fullcost} summarizes the episodic classification-head
benchmark. It uses synthetic 384-dimensional support and query embeddings
and randomly initialized heads, with the encoder replaced by an identity
module. Image loading and ViT feature extraction are excluded. Forward
latency is measured in evaluation mode without gradients. Forward + backward
latency (F+B) includes gradient clearing, a training-mode forward pass,
cross-entropy calculation and backpropagation, excluding the optimizer step.

The benchmark implementation uses float32 tensors without automatic mixed
precision, with default schedules of 20 warm-up calls and 100 repetitions
for inference, and five warm-up calls and 25 repetitions for F+B. CUDA
synchronization follows warm-up and each measured call. Device-specific
execution settings were not retained, so the timings describe this
implementation rather than hardware-independent performance.

Peak CUDA allocated memory is measured during inference and expressed in
MiB ($2^{20}$ bytes). It includes resident objects, including the support-only
head, and does not isolate activation memory or measure training memory.
The support-only implementation requests attention weights, whereas WIPT
does not; the measurements consequently compare these implementations rather
than uniformly optimized attention kernels.

\begin{table*}[!htb]
\centering
\scriptsize
\setlength{\tabcolsep}{4pt}
\caption{Head benchmark for one 5-way episode with 75 queries.
PN and SO are baseline rows; numeric $g$ denotes WIPT query-group size.
Time is in milliseconds; memory is in MiB. F+B includes a training-mode
forward pass, loss and backpropagation, without an optimizer step.
P95 is the 95th percentile, not a confidence interval. Dashes indicate
metrics outside the recorded benchmark coverage. Token-pair counts omit the common attention-head multiplier.}
\label{tab:fullcost}
\begin{tabular}{ccrrrrrr}
\toprule
Shot & $g$/head & Token pairs & Fwd median & Fwd P95 & F+B median & F+B P95 & Peak MiB \\
\midrule
1 & PN & 0 & 0.25 & 0.26 & -- & -- & -- \\
1 & SO & 50 & 1.25 & 1.36 & -- & -- & -- \\
1 & 1 & 5,400 & 1.00 & 1.19 & 5.54 & 5.88 & 44.66 \\
1 & 2 & 3,698 & 2.14 & 2.36 & 9.61 & 9.89 & 48.76 \\
1 & 3 & 3,200 & 1.27 & 1.40 & 5.72 & 6.33 & 48.08 \\
1 & 4 & 3,044 & 2.05 & 2.51 & 9.60 & 9.72 & 46.77 \\
1 & 5 & 3,000 & 1.18 & 1.39 & 5.67 & 5.95 & 46.53 \\
\addlinespace
5 & PN & 0 & 0.34 & 0.43 & -- & -- & -- \\
5 & SO & 1,250 & 3.45 & 3.52 & -- & -- & -- \\
5 & 1 & 101,400 & 2.40 & 2.80 & 8.43 & 10.41 & 75.79 \\
5 & 2 & 55,298 & 3.38 & 5.34 & 12.34 & 13.05 & 64.13 \\
5 & 3 & 39,200 & 1.65 & 2.38 & 6.75 & 11.64 & 58.89 \\
5 & 4 & 31,844 & 2.87 & 5.10 & 12.57 & 12.73 & 54.68 \\
5 & 5 & 27,000 & 3.09 & 3.50 & 6.90 & 7.19 & 53.58 \\
\bottomrule
\end{tabular}
\end{table*}

Table~\ref{tab:selection} summarizes the epoch and elapsed time at which
the best validation checkpoint was obtained. Every reported multi-query
accuracy condition has five independently trained heads (seeds 0--4), and
all five checkpoints are retained. The timing summary for 5-shot $g>1$
uses the four runs with exported selection-time records; this subset does
not limit the five-seed accuracy analysis. Elapsed times for $g=1$ were not
recorded. Time to the selected checkpoint differs from total training time
and is not a measure of convergence.

\begin{table}[!htb]
\centering
\footnotesize
\caption{WIPT validation-checkpoint selection. Epoch and elapsed minutes
are median [minimum, maximum] over the summarized runs. The third column
gives the number of runs contributing to this table, not the number of
trained models: all conditions have five trained models. Dashes indicate
unrecorded elapsed times.}
\label{tab:selection}
\begin{tabular}{ccccc}
\toprule
Shot & $g$ & Runs summarized & Best epoch & Minutes to best \\
\midrule
1 & 1 & 5 & 2 [1, 4] & -- \\
1 & 3 & 5 & 4 [1, 7] & 4.61 [1.16, 11.21] \\
1 & 5 & 5 & 4 [1, 7] & 4.62 [1.16, 8.08] \\
\addlinespace
5 & 1 & 5 & 3 [1, 25] & -- \\
5 & 2 & 4 & 2 [1, 49] & 2.70 [1.33, 67.29] \\
5 & 3 & 4 & 2 [1, 49] & 2.74 [1.35, 67.71] \\
5 & 4 & 4 & 2 [1, 49] & 2.74 [1.36, 66.90] \\
5 & 5 & 4 & 2 [1, 49] & 2.56 [1.26, 61.67] \\
\bottomrule
\end{tabular}
\end{table}

\FloatBarrier

\FloatBarrier
\section{Detailed related-work comparison}
\setcounter{figure}{0}
\setcounter{table}{0}
\label{app:related}

\subsection{Metric and embedding adaptation in few-shot learning}
Matching Networks \cite{vinyals2016matching}, ProtoNet
\cite{snell2017prototypical}, and Relation Networks \cite{sung2018relation}
classify queries using a small support set and a learned or fixed similarity
rule. MAML instead learns an initialization that can be rapidly optimized
\cite{finn2017maml}. FEAT uses a Transformer set-to-set function to contextualize support-derived
class representations within an episode \cite{feat2020}. Its standard
prototype-adaptation path does not include the current query in that set;
the query embedding is subsequently compared with the adapted representatives.
This description distinguishes class-representation adaptation from an
independent learned projection of each image. PrototypeFormer is also Transformer-based, but
its prototype extraction module combines a class prototype token with
within-class support embeddings; support-derived sub-prototypes provide a
prototype-contrastive training objective
\cite{prototypeformer}. WIPT differs from these support-centric approaches by
placing the current query in the same global-embedding sequence as the support
set and by making the resulting class means query-specific.

\subsection{Query-conditioned and transductive prototype adaptation}
Cross\-Transformers use one unlabelled query to find spatial correspondences with
labelled support images \cite{crosstransformer2020}. Concretely, attention
matches local query regions to relevant regions of support images, and the
aligned feature vectors are compared to score each class. This retains spatial
information that is absent from one global vector per image. QPN similarly argues that
one fixed class prototype need not fit all queries and constructs a
query-specific region-level prototype \cite{qpn2021}. HCPNet incorporates the
query feature into prototype formation and adds contrastive regularization for
remote-sensing few-shot classification \cite{hcpnet2023}. These works show that
single-query conditioning is viable, but they operate on spatial or richer
learned representations rather than isolating global frozen embeddings.

Other approaches deliberately use multiple target queries. FSL-PRS treats the
support and query sets as one global self-attention context and then adds
high-confidence pseudo-labelled queries to rectify prototypes
\cite{fslprs2024}. PRSN uses the query set to reconstruct prototypes within a
global/local architecture \cite{prsn2025}. RDProtoFusion refines naive
prototypes with query samples and combines this with prototype contrast and
multi-task fusion \cite{rdprotofusion2024}. APPL performs transductive
weighted-moving-average self-training on target queries \cite{appl2024}.
EfficientFSL combines lightweight ViT tuning with a Support-Query Attention
Block that moves prototypes toward the query-set distribution
\cite{efficientfsl2026}. WIPT intentionally omits those query-set and target
update mechanisms so that the effect of query participation can be measured in
a simpler inductive setting; Section~\ref{sec:multiquery_method} then relaxes
this restriction as a controlled comparison.

\begin{table*}[!htb]
\centering
\footnotesize
\setlength{\tabcolsep}{3pt}
\caption{Context and representation paths of the closest methods. ``Query
transformation'' means a change conditioned on the episode, not simply applying
a learned image encoder. This is a conceptual comparison; the custom
support-only control is not a reproduction of FEAT or PrototypeFormer.}
\label{tab:positioning}
\begin{tabularx}{\textwidth}{>{\raggedright\arraybackslash}p{2.15cm}>{\raggedright\arraybackslash}p{1.75cm}>{\raggedright\arraybackslash}X>{\raggedright\arraybackslash}X}
\toprule
Method & Context for adaptation & Support / prototype path & Query path and distinction \\
\midrule
FEAT \cite{feat2020} & Support set & Set-to-set adaptation of support-derived class representations & Query excluded from standard prototype-adaptation attention \\
Prototype\-Former \cite{prototypeformer} & Within-class supports & Prototype token and support embeddings; sub-prototype contrastive training & Query embedding compared with extracted prototypes; excluded from the extraction module \\
Cross\-Transformers \cite{crosstransformer2020} & One query + supports & Spatially aligned support features depend on the query & Local query descriptors guide support matching; spatial rather than joint global-token adaptation \\
QPN \cite{qpn2021} & One query + supports & Region-level query-specific prototypes & Local query representations guide hierarchical semantic matching \\
FSL-PRS \cite{fslprs2024} & Support + query set & Joint attention and pseudo-label prototype rectification & Query tokens participate in the shared attention context \\
APPL \cite{appl2024} & Support + query set & Parametric prototype learning and self-training & Multiple target queries support transductive self-training \\
\textbf{WIPT} & \textbf{One query + supports} & \textbf{Jointly transformed supports; query-specific class means} & \textbf{Query transformed with supports; no target-time parameter updates} \\
\bottomrule
\end{tabularx}
\end{table*}

\subsection{Cross-domain few-shot learning and source-domain variation}
CD-FSL introduces a distribution gap between source training and target
episodes. Feature-wise transformation (FWT) perturbs feature distributions
during source training to simulate unseen domains \cite{tseng2020fwt}, while
broader benchmarks demonstrate that performance depends strongly on the target
domain \cite{guo2020bscdfsl}. Task-specific adapters provide a lightweight
route to target adaptation \cite{taskadapters2022}. StyleAdv expands the source
style distribution using hard adversarial styles and shows that explicitly
training against source-domain appearance variation can improve CD-FSL
\cite{styleadv2023}. These results motivate our RQ4 extension: rather than
changing the frozen backbone, we expose only the episodic WIPT/support-only
heads to source pseudo-domains and test whether this improves natural
cross-domain transfer.

\FloatBarrier

\FloatBarrier
\section{Training and evaluation algorithm}
\setcounter{figure}{0}
\setcounter{table}{0}
\label{app:pipeline}
\begin{algorithm}[!htb]
\caption{Controlled training and cross-domain evaluation pipeline}
\label{alg:pipeline}
\small
\begin{algorithmic}[1]
\State Load pretrained ViT-S/16 encoder $f_\theta$ and freeze all $\theta$.
\For{shot count $K\in\{1,5\}$}
\For{training seed $r\in\{0,1,2,3,4\}$}
  \For{trainable head $h\in\{\text{support-only},\text{WIPT}\}$}
    \State Train $h$ episodically on miniImageNet train classes using frozen $f_\theta$.
    \State Select the checkpoint with the highest miniImageNet validation accuracy.
  \EndFor
\EndFor
\State ProtoNet receives no episodic training; it uses the same verified frozen encoder state.
\For{target domain $D\in\{\text{CUB},\text{EuroSAT},\text{ISIC}\}$}
  \For{evaluation seed $e\in\{42,123,2024,7,999\}$}
    \State Sample 300 matched 5-way episodes from $D$.
    \State Encode support and query images with frozen $f_\theta$.
    \State Evaluate ProtoNet and every trained head on the same episodes with no target update.
  \EndFor
\EndFor
\EndFor
\State Aggregate episode variation descriptively; summarize paired effects across training seeds.
\end{algorithmic}
\end{algorithm}

\FloatBarrier
\section{Multi-query protocol and complete accuracy results}
\setcounter{figure}{0}
\setcounter{table}{0}
\label{app:query}
\subsection{Multi-query context comparison}

To test whether the default single-query restriction sacrifices useful target
context, we generalize the WIPT sequence to a group of $g$ unlabelled queries,
\begin{equation}
\mathbf{T}^{(0)}_g =
[\,\mathbf{q}_1;\ldots;\mathbf{q}_g;\mathbf{s}_1;\ldots;\mathbf{s}_{NK}\,].
\end{equation}
The transformed supports are shared by the $g$ queries in that group, producing
one set of group-conditioned class prototypes; each transformed query is then
scored against those prototypes. Query labels are never used to form groups.
Because the episodic loader is naturally class-major, queries and labels are
shuffled together before grouping during training, validation and evaluation
so that $g>1$ does not receive artificial
same-class context. Any final incomplete group is processed at its natural
size. The $g=1$ path is exactly the original WIPT computation and remains
checkpoint-compatible. Complete groups are processed as one Transformer batch;
the possible remainder is processed separately. Scores from all groups are
concatenated before the episode-mean training loss is computed, so the remainder
is weighted by its number of queries.

For each tested shot count and group size, we use group-specific trained
heads and evaluate with the corresponding group size. The $g=1$ condition
reuses the standard WIPT checkpoints; $g>1$ is not merely an inference-time
regrouping of those weights. All conditions retain WIPT-2's architecture and
the five training seeds. We evaluate $g\in\{1,2,3,4,5\}$ in 5-shot and then
use $g\in\{1,3,5\}$ as a sparse follow-up in 1-shot, selected after inspecting
the 5-shot results. This follow-up is not a preregistered test of equivalence.
All five trained checkpoints are retained for each tested condition, including
5-shot $g=2,3,4,5$ and the 1-shot $g=3,5$ follow-up. Additional computational
cost results are reported in \ref{app:cost}.

\subsection{Computational cost}
WIPT-2 contains $3{,}549{,}696$ trainable head parameters, while the ViT-Small
backbone contains roughly 22 million frozen parameters. The support-only
Transformer has the same Transformer-head parameter count. For single-query
5-way 5-shot WIPT, each Transformer sequence contains 26 tokens. Grouping
queries increases tokens per sequence but amortizes the support tokens across
several queries. We therefore report both analytical attention-token pairs per
episode and head-only inference and forward + backward latency and CUDA peak memory in
Section~\ref{sec:multiquery_results} and \ref{app:cost}. For
$M=|Q|=75$ queries partitioned into groups with actual sizes $g_j$, the count is
\begin{equation}
C_{\mathrm{attn}}=L\sum_{j=1}^{\lceil M/g\rceil}(NK+g_j)^2.
\end{equation}
This convention includes all layers but omits the common six-head multiplier.
It counts token pairs, not FLOPs, and excludes projections and feed-forward
layers. The benchmark excludes image encoding and data loading; the shared ViT
cost is outside the measured head operation.

\subsection{RQ2: does additional query context improve accuracy?}

The full 5-shot sweep gives no evidence that jointly processing additional
unlabelled queries improves cross-domain accuracy. Relative to $g=1$, the mean
5-shot changes for $g=2$--5 are only $-0.012$ to $-0.017$ points on CUB,
$-0.459$ to $-0.502$ on EuroSAT and $-0.130$ to $-0.141$ on ISIC. Every
95\% training-seed interval includes zero. The 1-shot confirmation gives the
same qualitative conclusion: $g=3$ and $g=5$ are slightly positive on CUB
($+0.039$ and $+0.050$ points), slightly negative on EuroSAT and ISIC, and all
intervals include zero. Thus, across both shot regimes, we do not detect a reliable accuracy advantage from extra query context over independently processing one query.

This conclusion is not an artefact of class-major query ordering. Queries were
shuffled before grouping; in the 5-shot evaluation, groups of five contained
3.41 distinct classes on average and only $0.10\%$ were homogeneous. The same
post-hoc diversity check applies to the 1-shot $g=3$ and $g=5$ comparisons.

The computational trade-off is more nuanced. All $g$ values use exactly
$3{,}549{,}696$ trainable head parameters. In 5-shot, analytical attention work
falls from 101,400 token-pairs per episode at $g=1$ to 27,000 at $g=5$ (about a
73\% reduction) because support tokens are amortized across queries; recorded CUDA peak allocated memory also falls from 75.8 MiB to 53.6 MiB. In 1-shot, attention work
falls from 5,400 to 3,000 token-pairs, while recorded peak allocated memory is roughly
flat or slightly higher for $g>1$. Head-only forward latency is non-monotonic
rather than proportional to analytical attention work (for example, 5-shot
medians range from 1.65 to 3.38 ms for $g=2$--5 versus 2.40 ms for $g=1$), so
we do not claim that larger query groups are uniformly faster. At 5-shot, the recorded forward + backward median also changes
non-monotonically:
8.43 ms at $g=1$, 6.75 ms at $g=3$ and 6.90 ms at $g=5$.
\ref{app:cost} reports all inference and forward + backward medians and 95th
percentiles, peak memory, and descriptive checkpoint-selection times.
The practical advantage of $g=1$ is independent, streaming-compatible inference.
The accuracy intervals do not establish equivalence: for example, the 5-shot
EuroSAT $g=5$ difference is compatible with changes from $-1.536$ to $+0.536$
points. Checkpoint-selection times describe when the best validation model was
obtained; they do not establish a convergence-speed advantage.

\begin{table*}[!htb]
\centering
\small
\caption{Multi-query accuracy change relative to single-query WIPT ($g=1$), in
percentage points. Values are mean $\pm$ 95\% Student's $t$ interval across five
paired training seeds. Every interval includes zero. The 1-shot study uses the
sparse follow-up $g\in\{1,3,5\}$ chosen after the 5-shot sweep.}
\label{tab:multiquery}
\begin{tabular}{c c c c c}
\toprule
Shot & $g$ & CUB & EuroSAT & ISIC \\
\midrule
5 & 2 & $-0.015\pm0.074$ & $-0.484\pm0.983$ & $-0.141\pm0.470$ \\
5 & 3 & $-0.017\pm0.065$ & $-0.459\pm0.913$ & $-0.130\pm0.455$ \\
5 & 4 & $-0.015\pm0.060$ & $-0.502\pm0.989$ & $-0.135\pm0.460$ \\
5 & 5 & $-0.012\pm0.061$ & $-0.500\pm1.036$ & $-0.138\pm0.468$ \\
\addlinespace
1 & 3 & $+0.039\pm0.115$ & $-0.192\pm0.626$ & $-0.103\pm0.171$ \\
1 & 5 & $+0.050\pm0.127$ & $-0.159\pm0.634$ & $-0.092\pm0.166$ \\
\bottomrule
\end{tabular}
\end{table*}

\FloatBarrier

\FloatBarrier
\section{Detailed decision-change and boundary analysis}
\setcounter{figure}{0}
\setcounter{table}{0}
\label{app:boundary}
\subsection{Mechanism and boundary analysis}

To study why the effect changes with domain and shot count, we evaluate every
standard WIPT training seed on matched target episodes and record two levels of
diagnostics. At the episode/domain level, \emph{support uncertainty} compares
the support-derived class mean with the labelled query-class centroid,
normalized by mean inter-class query-centroid distance. Query labels are used
only after prediction for this analysis-only oracle quantity. We also measure
\emph{conditioning variability}, the normalized change in a class prototype
across different queries, together with prototype displacement, within-class
query scatter and ProtoNet margins. Correlations between individual episode
metrics and WIPT gain are treated as exploratory rather than causal evidence.

At the query level, we record whether WIPT rescues a ProtoNet error or breaks a
ProtoNet success. For each shot/domain combination, the same 1,500 episodes
yield 112,500 queries per trained head. Define the raw ProtoNet class mean
$\mathbf{p}^{\mathrm{PN}}_c=K^{-1}\sum_{i:y_i=c}\mathbf{s}_i$ and true-class margin
\begin{equation}
m_{\mathrm{PN}}(\mathbf q,y)=
\min_{c\ne y}\|\mathbf q-\mathbf p^{\mathrm{PN}}_c\|_2
-\|\mathbf q-\mathbf p^{\mathrm{PN}}_y\|_2.
\label{eq:margin}
\end{equation}
A negative margin is a ProtoNet error and a positive margin is a success;
there are no exact zero-margin ties among the evaluated queries. Thus a
signed WIPT--ProtoNet accuracy change is necessarily nonnegative for a
negative-margin query and nonpositive for a positive-margin query. That sign
pattern alone cannot establish a mechanism.

To examine where decisions change, we sort queries by raw margin separately
within ProtoNet errors and successes, then divide each subset into ten
approximately equal-count bins. No bin crosses zero; bin counts differ between
the two subsets. We plot the fraction of correctness transitions within each
bin, with a 95\% Student's $t$ interval across the five WIPT training seeds.
For an additional descriptive localization measure, we compute the fraction
of all rescues and breaks contained in the 20\% of queries with the smallest
$|m_{\mathrm{PN}}|$. This post-hoc summary is not a tuned prediction rule.
These raw margins are distinct from the scale-normalized geometry diagnostic
in \ref{app:exploratory}.

The aggregate accuracy effect can be decomposed exactly. If $a$ is ProtoNet
accuracy as a fraction, $r$ is the rescue fraction among its errors and $b$ is
the break fraction among its successes, then
Equation~\ref{eq:rescue_break} gives this decomposition.
We compute $r$ and $b$ from pooled query counts separately for each trained
head. Multiplying both terms by 100 gives comparable contributions per 100
queries, whose difference is the accuracy gain in percentage points. These
pooled rates differ from an unweighted average of episode-wise conditional
rates, particularly when some episodes contain very few ProtoNet errors.

\subsection{RQ3: where do the shot- and domain-dependent effects arise?}

More support examples reduce the observed uncertainty in the class estimate.
The normalized support-to-query centroid error approximately halves from
1-shot to 5-shot: CUB $0.758\to0.382$, EuroSAT $1.001\to0.503$ and
ISIC $2.062\to1.029$. Query-conditioned prototype variability also falls,
from $0.0145\to0.0034$, $0.0244\to0.0066$ and $0.0327\to0.0081$,
respectively. These are descriptive associations with shot count, rather than
evidence that either statistic alone causes an accuracy improvement.

The rescue/break decomposition explains the arithmetic of the net changes.
On 1-shot EuroSAT, WIPT rescues 16.42\% of ProtoNet errors and breaks 6.69\%
of its successes. Because these rates have different denominators, subtracting
them would not give an accuracy change. Expressed per 100 queries, the mean
contributions are 6.223 rescued and 4.155 broken, giving the observed
$+2.068$ points. In 5-shot EuroSAT, the corresponding contributions are
2.183 and 2.905, giving $-0.721$ points. The baseline error pool shrinks from
37.91\% to 17.24\% as the shot count increases.

CUB illustrates the effect of saturation: in 5-shot, only 2,385 of 112,500
queries are ProtoNet errors. WIPT rescues 0.178 predictions per 100 queries but
breaks 0.266, leaving a small negative net change despite rescuing 8.38\% of
baseline errors. In 1-shot ISIC, 1.879 rescues per 100 queries are outweighed by
2.098 breaks. Figure~\ref{fig:shotmechanism} puts all six conditions on the
same denominator and displays the support diagnostics separately.

\begin{figure*}[!htb]
\centering
\includegraphics[width=\textwidth]{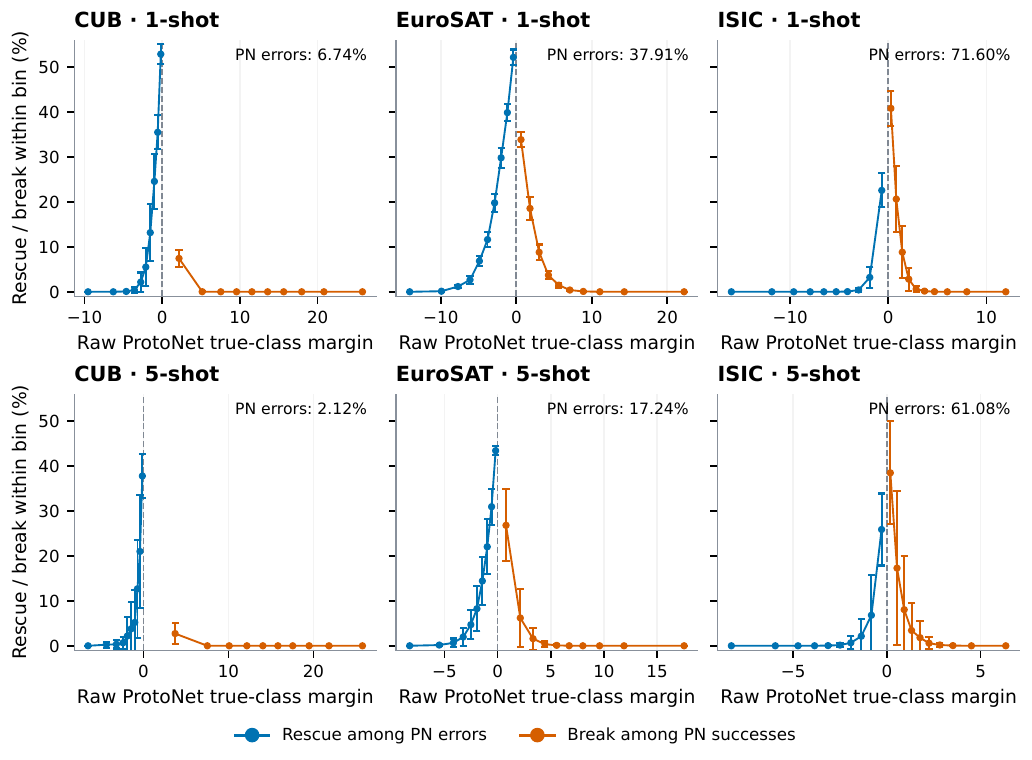}
\caption{Rescue and break probability versus raw ProtoNet true-class margin.
For each of the six shot/domain conditions, ProtoNet errors (blue) and
successes (orange) are separately sorted into ten approximately equal-count
bins. Points are plotted at the median margin and show rescue probability
among errors or break probability among successes. Error bars are unadjusted
95\% Student's $t$ intervals across five WIPT seeds. No bin crosses the
zero-margin boundary. Counts differ between the two sides; the panel
annotation gives the fraction of all queries that are ProtoNet errors.
Marginal accuracy contributions are shown separately in
Figure~\ref{fig:shotmechanism}.}
\label{fig:boundary_all}
\end{figure*}

The boundary analysis measures rescue and break probabilities without
mixing errors and successes in a bin. Figure~\ref{fig:boundary_all} shows that
rescue and break probabilities are largest near zero margin and decline away
from the boundary. This localization contains information beyond the
structural sign constraint in Equation~\ref{eq:margin}. The 20\% of queries
nearest the boundary contain, on average across training seeds, 100.0\%,
68.4\% and 96.5\% of rescues and breaks on 1-shot CUB, EuroSAT and ISIC;
the corresponding 5-shot fractions are 100.0\%, 91.1\% and 91.4\%.
These post-hoc concentrations quantify where WIPT changes baseline
correctness, but do not identify a causal pathway or guarantee a positive
change. The analysis concerns changes in correctness and excludes switches between
two incorrect predicted classes.

Episode-level associations between WIPT gain and individual geometry metrics
remain weak: across all per-seed Pearson and Spearman analyses, the largest
absolute correlation coefficient is $0.146$. We
therefore treat support uncertainty and prototype variability as exploratory
diagnostics, and the boundary concentration as a reproduced pattern of decision
changes. Neither establishes why a particular transformation rescues one query
but breaks another.

\FloatBarrier

\FloatBarrier
\section{Controlled corruption: protocol and full response}
\setcounter{figure}{0}
\setcounter{table}{0}
\label{app:corruption}
\subsection{Controlled corruption protocol}
\label{sec:corruption_protocol}
Natural cross-domain evaluation answers whether WIPT transfers to real target
domains; controlled corruptions answer a different question: how does the
relative ordering change as a known appearance perturbation is increased while
the target dataset is held fixed? We therefore use corruption only as a stress
test of the prototype-adaptation mechanism, not as a substitute for a natural
cross-domain benchmark.

Category A applies brightness followed by contrast; category B applies Gaussian
noise; category C applies brightness, contrast and then Gaussian noise. The
corruption is applied to both support and query images. Each category-severity
condition uses 200 episodes for each of the five evaluation seeds (1,000
episodes). The replicated analysis is concentrated on EuroSAT because the
initial diagnostic showed the strongest corruption interaction there. The
severity-0 clean reference uses the same 1,000-episode bank as the corruption
conditions, rather than the 1,500-episode bank in Table~\ref{tab:crossdomain};
small differences between these clean estimates are therefore expected.

\begin{table}[!htb]
\centering
\caption{Corruption schedule. Brightness is an additive RGB offset on the
$[0,1]$ scale, contrast is a multiplier around the per-channel image mean, and
Gaussian noise is zero-mean with standard deviation $\sigma$.}
\label{tab:corruption_schedule}
\begin{tabular}{cccc}
\toprule
Severity & Brightness offset & Contrast multiplier & Gaussian $\sigma$ \\
\midrule
1 & $+0.10$ & 0.75 & 0.08 \\
2 & $+0.20$ & 0.60 & 0.12 \\
3 & $+0.30$ & 0.50 & 0.18 \\
4 & $+0.40$ & 0.40 & 0.26 \\
5 & $+0.50$ & 0.30 & 0.38 \\
\bottomrule
\end{tabular}
\end{table}

\FloatBarrier
\subsection{Controlled corruption response}

The replicated EuroSAT corruption experiment further shows that ``adaptation''
should not be equated with uniform robustness. Relative to frozen ProtoNet,
structured brightness/contrast corruption mostly favours ProtoNet or produces a
tie. Gaussian noise remains close: the mean WIPT difference moves from
$-0.39$ points at severity 1 to $+0.26$ at severity 5. Combined corruption
shows the clearest reproducible crossover, from $-0.26$ points at severity 3 to
$+0.75$ at severity 4 and $+1.02$ at severity 5; all five WIPT runs are positive
at severities 4 and 5.

Relative to the matched support-only Transformer, WIPT is positive on average
in every EuroSAT corruption condition and is especially consistent under
stochastic and combined corruption. The comparison therefore matters: query
participation improves the robustness of a transformed-support head, but does
not guarantee superiority over the untransformed ProtoNet class mean. The
high-severity combined crossover is the part of the ProtoNet comparison that
survives independent retraining.

\begin{figure*}[!htb]
\centering
\includegraphics[width=\textwidth]{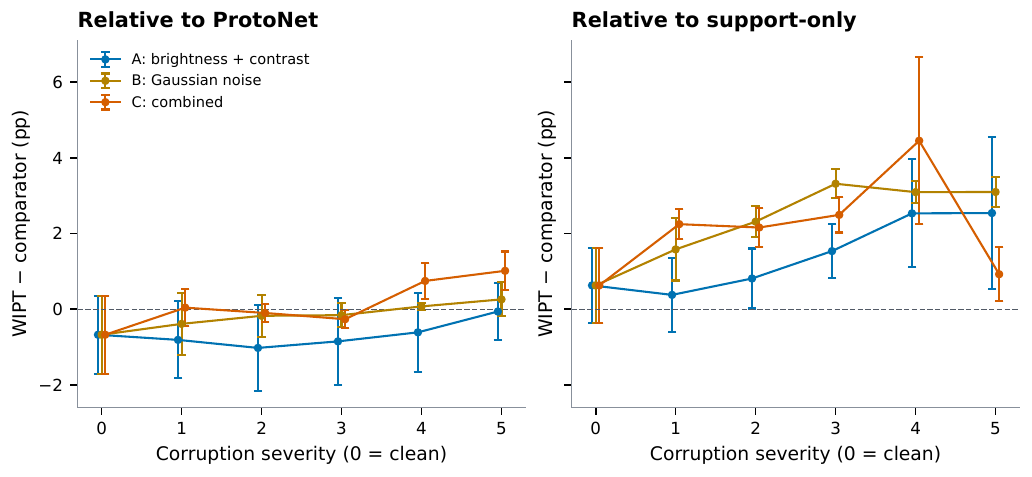}
\caption{Replicated EuroSAT corruption response across five independent 5-shot
training runs. Curves show mean WIPT accuracy difference under structured (A),
stochastic (B) and combined (C) corruption; error bars are 95\% training-seed
intervals. (a) Relative to frozen ProtoNet, effects are small and regime
dependent, with combined corruption crossing in favour of WIPT at severities 4
and 5. (b) Relative to the matched support-only control, WIPT remains positive
on average across the corruption study.}
\label{fig:corruption}
\end{figure*}

\FloatBarrier

\FloatBarrier
\section{Source-only shared-shift training: protocol and full results}
\setcounter{figure}{0}
\setcounter{table}{0}
\label{app:shift}
\subsection{Source-only shared-shift training extension}
\label{sec:shift_protocol}
RQ4 tests one feasible interpretation of explicitly training the episodic head
to tolerate domain variation. The encoder remains frozen and no target-domain
images are used. On each source training episode, with probability 0.5 we train
on the ordinary miniImageNet episode and with probability 0.5 we apply one
shared pseudo-domain appearance transform to both support and query images. A
shared transform is closest to the target evaluation setting, where support and
query belong to one target domain. The transform samples symmetric brightness
($\pm0.25$ at full strength), contrast ($0.60$--$1.40$), saturation
($0.55$--$1.45$), gamma ($0.70$--$1.40$), red/blue channel scaling
($0.85$--$1.15$), and Gaussian noise ($\sigma\le0.16$). Shift strength follows
a curriculum from 0.25 to 1.0 over 50 epochs. Checkpoint selection equally
weights clean and maximum-strength shifted miniImageNet validation accuracy.
Five independently trained shared-shift WIPT and shared-shift support-only heads
are evaluated in both 1-shot and 5-shot settings. As a development-only follow-up,
we also screen one \emph{mixed} training seed per shot regime; on shifted episodes,
this variant randomly uses either a shared support/query pseudo-domain or independent
support/query pseudo-domains. Because this follow-up has only one training seed, it
is reported descriptively in \ref{app:mixedshift} and is not used for a
replicated claim.

\subsection{RQ4: source-only shared-shift training}

Explicitly exposing the episodic head to shared source pseudo-domains provides
no evidence of improvement over standard WIPT on the natural target domains. In 1-shot, shared-shift
WIPT changes standard WIPT by $-0.063\pm0.099$ points on CUB,
$-0.399\pm1.254$ on EuroSAT and $-0.055\pm0.302$ on ISIC, where the $\pm$
term is the paired 95\% training-seed interval. In 5-shot, the corresponding
changes are $+0.018\pm0.144$, $-0.419\pm1.285$ and $-0.198\pm0.655$ points.
Every interval includes zero. This feasible source-only curriculum therefore
provides no evidence of improvement over the ordinary WIPT training protocol.

The result does not imply that query participation disappears under shift-aware
training. Against the \emph{shift-trained} support-only control, WIPT remains
higher by $+1.72$ points on 1-shot EuroSAT on average and, in 5-shot, by
$+0.82$ on EuroSAT and $+1.47$ on ISIC; all five 5-shot runs are positive in
the latter two domains. The negative RQ4 result is specifically about whether
this pseudo-domain curriculum improves \emph{standard WIPT}, not about whether
the query is useful relative to a matched transformed-support head.

\begin{table*}[!htb]
\centering
\footnotesize
\caption{Source-only shared-shift extension. Standard WIPT and shared-shift
WIPT are means across five training seeds. $\Delta$ is shared-shift WIPT minus
standard WIPT with a paired 95\% training-seed interval. No interval excludes
zero.}
\label{tab:shift}
\begin{tabular}{c l c c c}
\toprule
Shot & Domain & Standard WIPT & Shared-shift WIPT & $\Delta$ shift--standard (pp) \\
\midrule
1 & CUB & 93.465 & 93.402 & $-0.063\pm0.099$ \\
1 & EuroSAT & 64.159 & 63.761 & $-0.399\pm1.254$ \\
1 & ISIC & 28.177 & 28.122 & $-0.055\pm0.302$ \\
\addlinespace
5 & CUB & 97.791 & 97.809 & $+0.018\pm0.144$ \\
5 & EuroSAT & 82.041 & 81.623 & $-0.419\pm1.285$ \\
5 & ISIC & 38.380 & 38.182 & $-0.198\pm0.655$ \\
\bottomrule
\end{tabular}
\end{table*}

\FloatBarrier

\section*{CRediT authorship contribution statement}
\textbf{Rushab Rasik Karania:} Conceptualization, Methodology, Software, Investigation, Visualization, Writing.\par
\textbf{Tomas Maul:} Conceptualization, Methodology, Supervision, Review.

\section*{Declaration of competing interest}
The authors declare that they have no known competing financial interests or
personal relationships that could have appeared to influence the work reported
in this paper.

\section*{Funding}
This research did not receive any specific grant from funding agencies in the
public, commercial, or not-for-profit sectors.

\section*{Data and code availability}
The experiments use miniImageNet, CUB-200-2011, EuroSAT and ISIC 2019,
with the partitions and episode construction described in Section~\ref{sec:setup}.
The implementation, trained-head checkpoints and experimental result records
are retained by the authors. A public code repository is not provided with
this manuscript version.

\end{document}